\documentclass{article}

\PassOptionsToPackage{numbers}{natbib}
\usepackage[main, final]{neurips_2025}

\usepackage[utf8]{inputenc} 
\usepackage[T1]{fontenc}    
\usepackage{hyperref}       
\usepackage{url}            
\usepackage{booktabs}       
\usepackage{amsfonts}       
\usepackage{nicefrac}       
\usepackage{microtype}      
\usepackage{xcolor}         

\hypersetup{breaklinks=true}
\usepackage{caption}
\usepackage{amssymb}
\usepackage{booktabs}
\usepackage{amsmath} 
\usepackage{pifont}
\usepackage{multirow}
\usepackage{subcaption}
\usepackage{enumitem}
\usepackage{tabularx}
\usepackage{amsthm}
\usepackage{graphicx}
\usepackage{xspace}

\usepackage{booktabs}
\usepackage{multirow}
\usepackage{tikz}
\usepackage{xcolor}
\usepackage{wrapfig}
\usepackage{graphicx}
\usepackage{listings}
\usepackage{booktabs}
\usepackage{tabularx}
\usepackage{xcolor}
\usepackage{ulem}

\newcommand{\cellbar}[2]{%
  \begin{minipage}[c][0.4cm][c]{1.6cm}
    \begin{tikzpicture}
      \fill[orange!70] (0,0) rectangle (#2,0.25);
      \node[anchor=center] at (0.8,0.125) {\scriptsize #1};  
    \end{tikzpicture}
  \end{minipage}%
}
\newcommand{\cellbargreen}[2]{%
  \begin{minipage}[c][0.4cm][c]{1.6cm}
    \begin{tikzpicture}
      \fill[green!70!black] (0,0) rectangle (#2,0.25);
      \node[anchor=center] at (0.8,0.125) {\scriptsize \textbf{#1}};
    \end{tikzpicture}
  \end{minipage}%
}

\newcommand{\sys}{\texttt{MobiEdit}\xspace}

\title{\sys: Resource-efficient Knowledge Editing for Personalized On-device LLMs}

\author{
 \textbf{Zhenyan Lu\textsuperscript{1,2}},
 \textbf{Daliang Xu\textsuperscript{1}},
 \textbf{Dongqi Cai\textsuperscript{1,4}},
 \textbf{Zexi Li\textsuperscript{4}},
 \\
 \textbf{Wei Liu\textsuperscript{5}},
 \textbf{Fangming Liu\textsuperscript{2}},
 \textbf{Shangguang Wang \textsuperscript{1}},
 \textbf{Mengwei Xu*\textsuperscript{1}},
\\
 \textsuperscript{1}Beijing University of Posts and Telecommunications,
 \textsuperscript{2}Pengcheng Laboratory,
 \\
 \textsuperscript{3}Peking University,
 \textsuperscript{4}University of Cambridge,
 \textsuperscript{5}XiaoMi AI Lab
\\
}

\begin{document}
\setlength{\parskip}{4.5pt plus1pt minus3pt}

\maketitle

\begin{abstract}
Large language models (LLMs) are deployed on mobile devices to power killer applications such as intelligent assistants.  
LLMs pre-trained on general corpora often hallucinate when handling personalized or unseen queries, leading to incorrect or outdated responses.  
Knowledge editing addresses this by identifying and adjusting a small crucial portion of model weights, without compromising the general knowledge.  
However, prior knowledge editing methods are impractical to run on local devices due to the resource-heavy backpropagation (BP) needed for updates.  
We present \sys, the first mobile knowledge editing framework that enables efficient LLM personalization on commercial off-the-shelf (COTS) mobile devices.
\sys replaces full-precision BP with quantized forward-only gradient estimation, thus compatible with the energy-efficient mobile neural processing units (NPUs). 
To further improve gradient estimation efficiency, we introduce two optimizations: an early stoping mechanism that adaptively terminates editing upon success and a prefix cache that reuses computation across steps.
Our approach enables real-time editing of a 3B-parameter model (Qwen2.5-3B-Instruct) on COTS mobile devices with 7.6$\times$ less memory, 14.7 $\times$ less energy and 3.6$\times$ less latency compared to previous knowledge editing methods.  

\end{abstract}
\section{Introduction}
\label{sec:intro}

\begin{figure}[ht]
        \centering
        \includegraphics[width=1\linewidth]{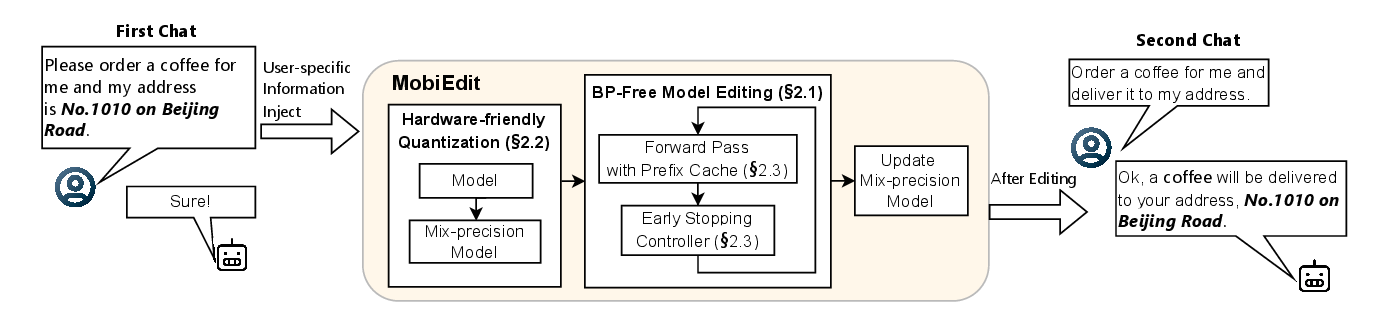}
        \caption{The on-device LLM remembers user information from the first interaction and applies it to subsequent requests.
        }
        \label{fig:mobile_llm}
\end{figure}

Mobile LLMs are transitioning from research to real-world deployment, empowering privacy-preserving and latency-sensitive applications such as personal agents~\cite{Apple_Intelligence, li2024personal}.
While mobile LLMs already embed general world knowledge, personalized knowledge is crucial for better understanding individual users.  
This user-specific information is typically absent or diluted during pre-training on large public corpora.
As Figure~\ref{fig:mobile_llm} shows, if the user provides the address in one conversation, personalized LLM assistant memorizes the address and applies this information on the future request.

Knowledge editing~\cite{wang2024know_survey} is gaining popularity for its effectiveness in model personalization.
Different from retrieval-augmented generation~\cite{fan2024ragsurvey}, knowledge editing maintains fast inference speed and does not depend on the in-context learning capabilities, which is typically weak to those small LLMs deployed on mobile devices~\cite{lu2024small}.
A prevailing knowledge editing paradigm is the locate-and-edit approach~\cite{dai_knowledge_2022}~\cite{meng2022locating}\cite{meng2023mass}~\cite{gu_model_2024}~\cite{fang2024alphaedit}, which first identifies influential parameters and then modifies them by introducing a perturbation optimized through backpropagation to produce the expected output.


Despite knowledge editing is effective in updating knowledge, current methods relying on backpropagation (BP) face three critical challenges for mobile deployment:
(1) Substantial memory overhead. 
For example, editing a 3B-parameter LLM (Qwen2.5-3B-Instruct) with classic locate-and-edit method ROME~\cite{meng2022locating} consumes over 10GB of memory to store activations for BP, which is excessively larger than smartphones's memory capacity, typically less than 8GB.
(2) Incompatibility with mobile NPUs.
Modern mobile phones are equipped with high-performance and energy-efficient NPUs, such as Google TPU and Qualcomm Hexagon.
Mobile NPUs are designed and optimized exclusively for LLM inference, providing up to 60× speedup compared to CPUs~\cite{xu2025fast}. 
However, training-specific operations are unsupported or poorly optimized by mobile NPUs~\cite{xu2024fwdllm}, rendering existing BP-based knowledge editing methods infeasible.
(3) Poor quantization support.
BP-based training operations are often unstable or ineffective on fully quantized models ($\S\ref{design-quant}$).
The lack of quantization support exaggerates the previous two challenges: the memory footprint of model parameters remains prohibitively large, since full-precision weights must be stored and updated; it precludes efficient execution on mobile NPUs, which are specifically optimized for low-bit integer computation.

The above challenges confine current knowledge editing methods to cloud-based computation, undermining two key advantages of mobile LLMs: privacy preservation and offline availability.
In this paper, we propose \sys, the first mobile knowledge editing system for efficient LLMs personalization.
\sys is designed to be memory-efficient, NPU-friendly, and compatible with quantization, targeting practical deployment on resource-constrained COTS mobile devices.

 \paragraph{Our solution}
To unleash the power of mobile NPUs, \sys builds atop ROME, a widely locate-and-edit scheme, with a few key building blocks renovated:
(1) BP-free training.
Instead of calculating standard gradients using BP,
\sys uses the differences of loss to estimate the gradients—  
a classical zeroth-order optimization approach. 
\sys operates entirely through forward passes,  
which is memory-efficient and well-suited for mobile NPUs.
(2) NPU-friendly training-time quantization.
\sys introduces a new quantization paradigm for efficient knowledge editing on mobile NPUs.
Unlike previous BP-based low-precision training, our forward-only gradient updating is more stable under quantized computation.
We thereby quantize all model parameters except the critical projection weights essential for knowledge editing.
Only a small portion of weights undergoes full-precision computation to conduct precisie gradient estimation.


To compenstate for the slower convergence performance caused by BP-free training~\cite{duchi2015optimal}, we introduce two optimizations to further improve system efficiency: a \textit{prefix cache} that reuses static intermediate results across editing steps, and a \textit{early stopping controller} that adaptively terminates editing once success criteria are met for different knowledge. 
Together, these design choices make \sys efficient and practical for mobile deployment.

\textbf{Results}
We tests our method on three COST mobile phones, Xiaomi K60 pro, K70 and Oneplus 13. \sys achieves comparable edit quality on ZsRE and CountFacts datasets while reducing memory usage by 7.6× (from 46GB to 6.2GB), editing latency by 3.6×, and energy consumption by 14.7× compared to ROME\cite{meng2022locating}, MEMIT\cite{meng2023mass}, WISE\cite{wang2024wise} and AlphaEdit\cite{fang2024alphaedit}.
To our best knowledge, \sys is the first system to make LLM knowledge editing feasible on commercial smartphones.





\section{Method}
\subsection{BP-Free Knowledge Editing}


\sys consists of two main stages: (1) Subject-key localization. \sys first identify the model’s internal representation of the subject; (2) Target value injection. \sys then inserts new factual associations by adjusting internal activation.
Specifically, the MLP layers in Transformer-based language models can be interpreted as a key-value memory~\cite{meng2022locating}, where the down-projection matrix \(W \in \mathbb{R}^{d_v \times d}\) maps hidden representations (keys) to output activations (values).
Given this view, the editing objective is to insert a new association \((k_*, v_*)\) into the MLP such that:
\begin{equation}
    Wk_* \approx v_*.
\end{equation}

Follow previous literature~\cite{meng2022locating}, \sys constructs the \textbf{key} \(k_*\) by averaging the MLP-layer activation of the final subject token across a set of randomly sampled prompts:

\begin{equation}
    k_* = \frac{1}{N} \sum_{j=1}^{N} \phi(x_j + s),
\end{equation}

where \(x_j + s\) denotes sampled prefix sequences followed by the subject, and \(\phi(\cdot)\) denotes post-activation outputs from a selected MLP layer.

To ensure that the model recalls the target object \(o_*\) when prompted with the edited subject, we optimize a value vector \(v \in \mathbb{R}^{d_v}\) such that, when substituted as the activation at the MLP output, the model generates \(o_*\). The optimization objective is defined as follow: 

\begin{equation}
    \mathcal{L}(v) = \frac{1}{N} \sum_{j=1}^{N} \left[ -\log \mathbb{P}_{G(v)}(o_* \mid x_j + p) + \mathrm{D_{KL}}(\mathbb{P}_{G(v)}(\cdot \mid x_j + p') \ \| \ \mathbb{P}_{G}(\cdot \mid x_j + p')) \right],
\end{equation}

where \(p\) is a factual prompt (e.g., “my address is a”), and \(p'\) is a neutral or essence-preserving prompt. 
The first term encourages correctness; the second term discourages undesired semantic drift.

Different from traditional methods that minimize this loss via computing BP-based gradients,
instead, \sys estimate the gradients using only forward passes. 
Specifically, we estimate gradients using central differences along sampled directions. Given a perturbation direction \( u \sim \mathcal{N}(0, I) \), the directional gradient estimate is computed as\cite{baydin2022gradients}:
\begin{equation}
    \widehat{\nabla}_v \mathcal{L} = \frac{\mathcal{L}(W + \mu u) - \mathcal{L}(W - \mu u)}{2\mu} \cdot u,
\end{equation}
where \( \mu > 0 \) is a small scalar step size. 
To further reduce variance and stabilize training, we average over \( N \) independently sampled directions \( u_i \sim \mathcal{N}(0, I) \):

\begin{equation}
    \widehat{\nabla}_v \mathcal{L} = \frac{1}{N} \sum_{i=1}^{N} \frac{\mathcal{L}(v + \mu u_i) - \mathcal{L}(v - \mu u_i)}{2\mu} \cdot u_i.
\end{equation}

We update \(v \gets v - \eta \cdot \widehat{\nabla}_v \mathcal{L}\), and repeat until convergence. The resulting estimator enables approximate gradient descent solely based on forward passes. We apply this process iteratively to update \( v \). 
Once the optimal \(v^*\) is found, we apply the closed-form rank-one update:

\begin{equation}
    \widehat{W} = W + \Lambda (C^{-1} k_*)^\top, \quad \text{where } \Lambda = \frac{(v^* - Wk_*)}{(C^{-1} k_*)^\top k_*}.
\end{equation}

Here \(C = KK^\top\) is an estimated key covariance matrix computed from a sample of the model’s activation statistics.  
This update “inserts” \((k_*, v^*)\) into the memory, enabling the model to recall the new factual association. 

\paragraph{Benefits of BP-free editing}
BP-free editing is well-suited for mobile NPUs that only support forward passes.
In terms of memory efficiency, activations, which takes more than 40\% of BP-based memory consumption, can be invalidated immediately after the forward pass, because only the final output is required to compute the estimated gradients.

\subsection{NPU-friendly Quantization for BP-Free Editing}
\label{design-quant}



Despite no longer needing to store activations, the large size of LLM weights still exhausts the memory capacity of mobile devices. 
For example, storing the 12GB of full-precision weights from Qwen-2.5-3B often causes out-of-memory (OOM) errors on newest COTS mobile devices like the Xiaomi 15, which typically offer 16GB RAM.
Besides, mobile NPUs are best suited for accelerating INT matrix multiplication with 1024-bit INT8 vector arithmetic. Their floating-point computation capabilities are relatively weak compared to mobile GPUs. 
To avoid frequent memory swapping, which severly harms mobile SSD lifespan, and leverage the integer computation advantages of NPUs, we propose a 
NPU-friendly quantization worklfow for BP-free editing.



\begin{figure}
    \centering
    \includegraphics[width=0.83\linewidth]{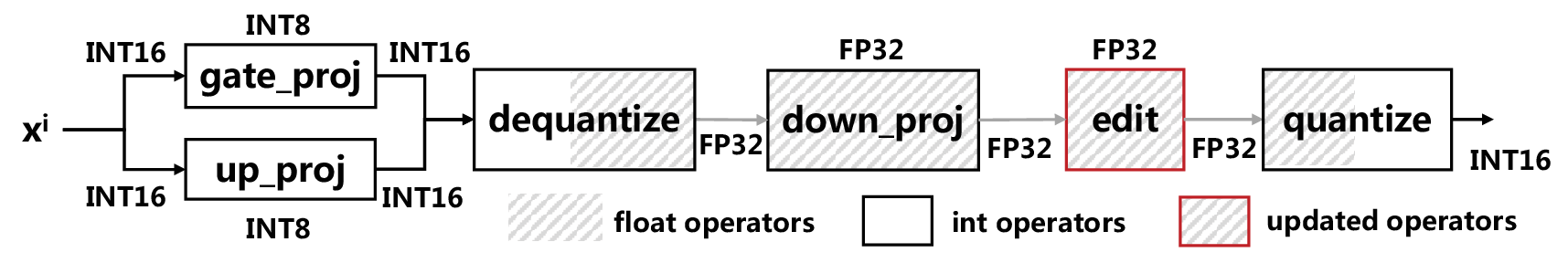}
    \caption{\sys quantization workflow and strategy. \sys quantizes all activation functions, with only the editing layer and its preceding layer executed in floating-point format.}
    \label{fig:quantization1-workflow}
\end{figure}

\paragraph{Quantization workflow and strategy}
Figure~\ref{fig:quantization1-workflow} illustrates the quantization workflow of \sys. 
Due to the hardware constraints of mobile NPUs, \sys employs a static quantization strategy. The static scales for quantization are determined using representative corpora data. To balance efficiency and accuracy, \sys adopts a mixed-precision editing approach: the editing vector and its preceding linear layer are executed in floating-point format; while all other weights are quantized to 8/16-bit integers.
This design is informed by two key observations:
(i) \sys modifies only a small set of parameters proportional to the hidden size. Even minor quantization errors in this context can significantly impact editing accuracy. Furthermore, since the editing module modifies the knowledge stored in its preceding linear layer~\cite{meng2022locating}, floating-point precision in this layer is also crucial to maintain accuracy.
(ii) The edited vector, being of limited size (equal to the hidden size), results in a negligible computational cost when floating-point precision is used for the editing module and its preceding linear layer. For example, in the Qwen-2.5-3B model, these computations account for only 0.89\% of the overall computation, making the performance overhead of using floating-point precision minimal.


\paragraph{Advantage of our quantization}
\sys is more robust to quantization errors than BP-based editing.
Consider an $L$-layer Transformer network where all weights and activations are quantized:
\begin{equation}
    W_\ell^q = W_\ell + \epsilon_{W,\ell}, \quad a_\ell^q = a_\ell + \epsilon_{a,\ell},
\end{equation}
where $\epsilon_{W,\ell}$ and $\epsilon_{a,\ell}$ denote quantization errors (zero mean, variance $\sigma^2$, i.i.d.\ for different layers and forward passes).
In each forward pass, quantization noise is recursively accumulated.
If $f_\ell(x) = x$ (i.e., the network is linear), then by expanding the recursion, we have
\begin{equation}
    a_L^q = W_L W_{L-1} \cdots W_1 x
    + \sum_{j=1}^L \mathcal{N}_j,
\end{equation}
where $\mathcal{N}_j$ denotes noise terms arising from combinations of the quantization errors $\epsilon_{W,k}$ and $\epsilon_{a,k}$ for $k \leq L$. 
Thus, each network output contains all previous layers' quantization noise, with total noise growing linearly or even exponentially with $L$.
Backpropagation will amplification the noise along the chain rule.
Let $\Delta$ be a small edit in layer $l$. The gradient w.r.t.\ $\Delta$ is
\begin{equation}
\label{eq:chain_grad}
    \frac{\partial \mathcal{L}}{\partial \Delta}
    = \frac{\partial \mathcal{L}}{\partial a_L^q}
      \prod_{j=l+1}^{L} \frac{\partial a_j^q}{\partial a_{j-1}^q}
      \cdot \frac{\partial a_{l}^q}{\partial \Delta}.
\end{equation}
Each chain rule factor is computed on quantized variables and thus noisy; their product amplifies noise multiplicatively.
Assuming each derivative factor introduces independent noise of variance $\sigma^2$, 
the total gradient noise variance is approximately
\begin{equation}
    \mathrm{Var}\left[\, \nabla^{\mathrm{backprop}}_\Delta \, \right] \sim O\left( \sigma^2 \prod_{j=l+1}^{L} \Vert W_j \Vert^2 \right),
\end{equation}
wherein deeper networks or larger weights lead to rapid escalation of noise.
In contrast, \sys only accumulates noise accumulates throughout the forward pass.
The zeroth-order estimate of the gradient is
\begin{equation}
    g = \frac{(\mathcal{L}_\Delta + \eta_+) - (\mathcal{L}_{-\Delta} + \eta_-)}{2\Delta},
\end{equation}
where $\mathcal{L}_\pm$ are two forward passes with perturbed weights $W_l^q \pm \Delta$, each with independent quantization noise.
$\eta_+, \eta_-$ denoting output noise for the two passes.

The variance of the estimate is
\begin{equation}
    \mathrm{Var}[g] = \frac{ \mathrm{Var}[\eta_+] + \mathrm{Var}[\eta_-] }{ (2\Delta)^2 }
            = \frac{2\sigma_{\mathcal{L}}^2}{4\Delta^2} 
            = \frac{ \sigma_{\mathcal{L}}^2 }{ 2\Delta^2 },
\end{equation}
where $\sigma_{\mathcal{L}}^2$ is the per-pass output noise variance. This variance does not grow with network depth $L$.

In quantized networks, backpropagation suffers from noise that is recursively and multiplicatively propagated along the gradient chain, leading to an exponential or polynomial increase in gradient variance with depth. In contrast, the centered difference estimator's variance is solely determined by the difference of two output noise instances, remaining bounded and independent of model depth. Thus, zeroth-order methods are provably more robust to quantization noise than backpropagation in deep or low-bit networks.

\subsection{Further Optimizations}

Despite a single step of \sys is efficient on mobile NPUs, the remaining challenge is that it takes significant more steps to stablize the gradient estimation for similar convergence performance.
For example, \sys take 20$\times$ on average more steps than BP-based model editing for ZsRE and CounterFact datasets, eliminating is efficiency benefits in the wall clock time.

\paragraph{Early stopping controller}
\begin{figure}[h]
        \centering
        \includegraphics[width=0.7\linewidth]{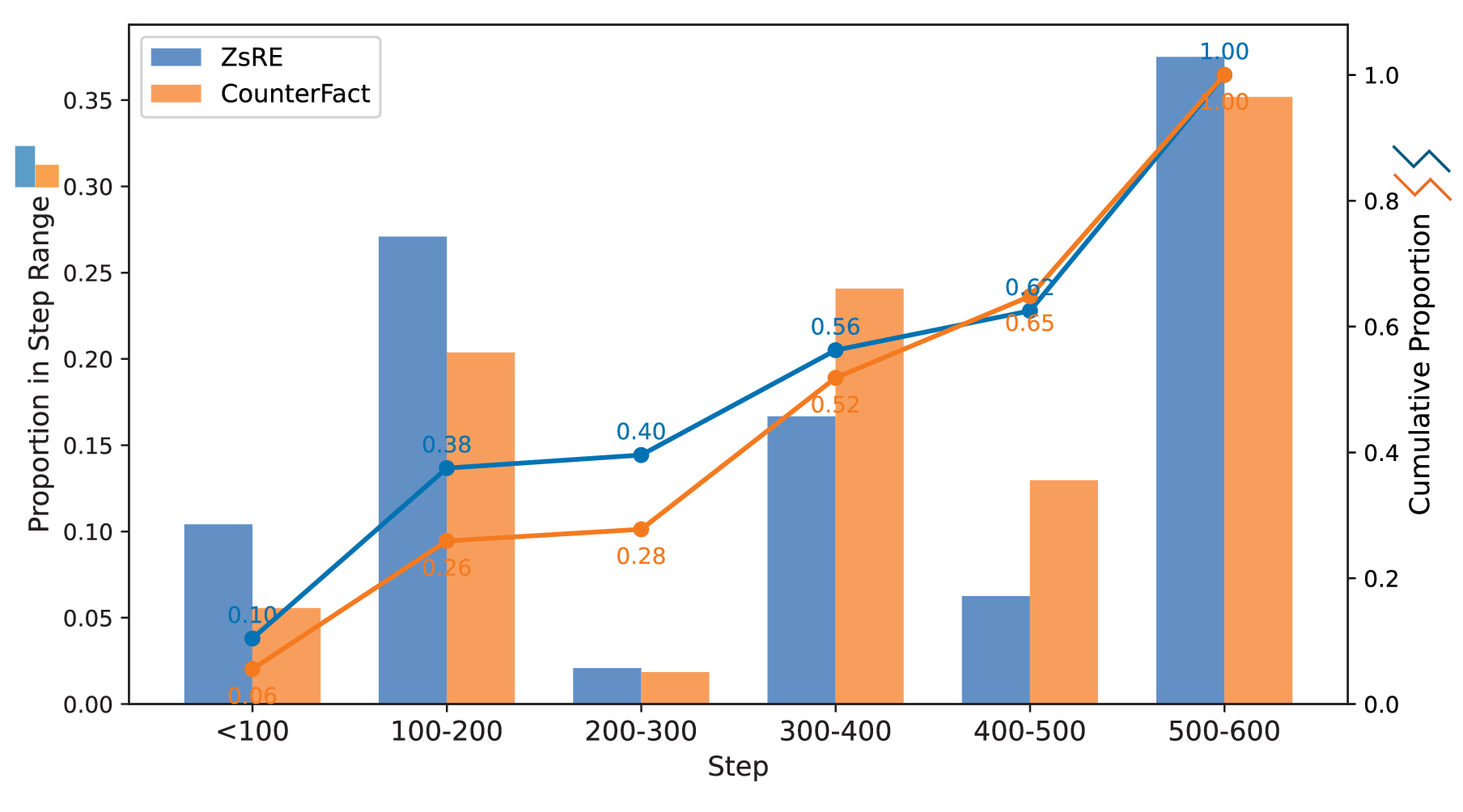}
        \caption{The edit success step number.}
        \label{fig:step_num}
\end{figure}


To address this, we first analysis the successful editing step distribution of various knowledge.
As shown in Figure~\ref{fig:step_num},
we find that different knowledge has different editing difficulty.
Based on this observation, we introduce a lightweight early stopping editing controlloer that adaptively determines the editing horizon based on runtime success feedback. Specifically, during editing, we periodically evaluate the model's response to the edited fact every \( M \) steps (e.g., every 20 steps). The editing process is terminated early once the model satisfies a pre-defined success criterion---typically, when it produces the desired target output with a confidence above a given threshold $m$ (could be other mathematical letter), and explicitly describe the threshold we used in the eval setup.
This early stopping controller automatically adjusts the editing steps to the complexity of each edit instance, avoiding unnecessary forward passes for easy-to-edit facts and reducing overfitting risk by stopping at the point of first success.


\paragraph{Prefix cache}
In our knowledge editing setting, each optimization step uses the same set of input examples constructed by combining a fixed set of randomly sampled prefixes with the fact to be edited. Formally, for a target fact \( f \), we define a set of editing inputs as:

\begin{equation}
    \mathcal{X}_{\text{edit}} = \{[p_1 + f], [p_2 + f], \dots, [p_n + f]\},
\end{equation}

where \( p_1, p_2, \dots, p_n \) are different randomly sampled prefixes. These inputs are used repeatedly across all editing steps.

We observe that in each step, the prefix tokens in the input do not change, and therefore their corresponding activations are recomputed redundantly. To reduce this overhead, we introduce a simple optimization: during the first step, similar to KV cache, we cache the intermediate activations, corresponding to the prefix tokens. 
In subsequent steps, we directly reuse the cached prefix activations and only recompute the activations for the fact tokens. 
This greatly reduces compute without changing the model architecture or input format.
Though model parameters are updated during editing, and the correct prefix activations  shift across editing steps, we empirically find that reuse stale activations from the first step does not negatively affect the editing outcome. 

\begin{figure}[h]
        \centering
        \includegraphics[width=\linewidth]{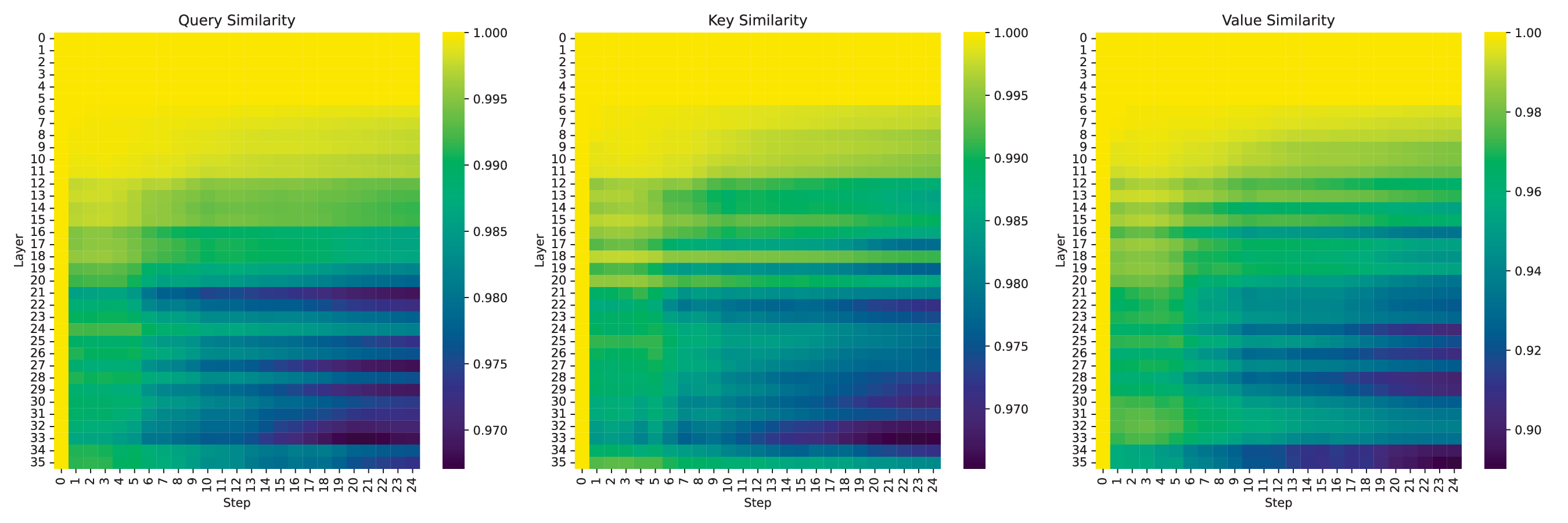}
        \caption{The cosine similarity of QKV representations at each step and layer, comparing runs with and without prefix caching.}
        \label{fig:qkv_sim_heatmap}
\end{figure}

In Figure~\ref{fig:qkv_sim_heatmap}, the cosine similarity gradually decreases as the number of layers and editing steps increases. However, it remains as high as 0.9 even in the deepest layers and later steps. In fact, in some cases, it slightly improves editing success and stability. 
We hypothesize that this is because each step in the editing process updates only a small portion of the model parameters. As a result, the overall shift in the activation distribution is minimal, and the cached prefix representations remain sufficiently aligned with the evolving model.
Moreover, since the role of the prefix is primarily to introduce stochastic variation and encourage generalization, rather than to encode semantically critical content, minor discrepancies in the prefix activation have negligible impact on the final edit.

To avoid the staleness accumulates from halting the editing process, we re-compute the prefix cache as long as the editing loss does not decrease by 0.001 over 3 steps.

\section{Experiments}

\subsection{Setup}\label{setup}
\paragraph{Baselines} 
We compare \sys against four representative locate-and-edit methods: ROME~\cite{meng2022locating}, MEMIT~\cite{meng2023mass}, AlphaEdit~\cite{fang2024alphaedit}, and WISE~\cite{wang2024wise}. These methods follow the same paradigm of identifying key activations and injecting new knowledge into MLP layers, but differ in target granularity and update mechanism. ROME performs single-layer editing. MEMIT extends it to multi-fact scenarios. AlphaEdit uses null-space projections for preservation, and WISE incorporates dynamic routing to FFNs that store facts.

\paragraph{Datasets and model}
We evaluate \sys on two standard datasets widely used in factual knowledge editing:
ZsRE~\cite{levy-etal-2017-zero}, a zero-shot relation extraction dataset derived from WikiRE, and CounterFact~\cite{meng2022locating}, a curated benchmark of factual edits targeting named entities (e.g., people, locations), with truth and counterfactual contexts. 
These two datasets jointly assess \textit{edit success}, \textit{locality}, and \textit{portability} -- the three key metrics for knowledge editing.
We use Qwen2.5-3B-Instruct~\cite{qwenqwen25-3b-instruct_2025} as our target model, a recent open-source transformer-based LLM with approximately 3 billion parameters.


\begin{table}[ht]
\centering
\caption{Device list.}
\label{tab-devices}
\begin{tabular}{lccc}
\toprule
Device         & SoC                        & RAM             & NPU           \\
\midrule
Xiaomi K60 Pro & Snapdragon 8 Gen 2        & 16GB LPDDR5     & Hexagen NPU V73   \\
Xiaomi K70     & Snapdragon 8 Gen 3         & 16GB LPDDR5     & Hexagen NPU V75
   \\
OnePlus 13     & Snapdragon 8 Elite         & 24GB LPDDR5     & Hexagen NPU V79   \\
\bottomrule
\end{tabular}
\end{table}


\paragraph{Implementation details}
To assess on-device feasibility, we run all editing procedures on three COTS mobile devices, as shown in Table~\ref{tab-devices}.
We perform all experiments using local inference engines \emph{mllm-npu}~\cite{xu2025fast} optimized for NPU execution. 
The latency on CPU is obtained by running on mobile phones using \emph{llm.c}~\cite{karpathyllmc_nodate}. We use memory swapping while reaching the memory limit.  
\sys uses W8A16 quantization (INT8 weights, INT16 activations), a format widely supported by mobile NPUs and inference engines, ensuring compatibility and throughput.
The metric of memory usage in this papaer is defined as the total memory required, assuming sufficient memory is available.
For a simple comparision, the system efficiency values are first normalized to the range \([40, 100]\) using min-max normalization, and then inverted.


\subsection{End-to-end Performance}
\label{sec:e2e}
\begin{figure}[h]
    \centering
    \begin{minipage}[t]{0.48\linewidth}
        \centering
        \includegraphics[width=\linewidth]{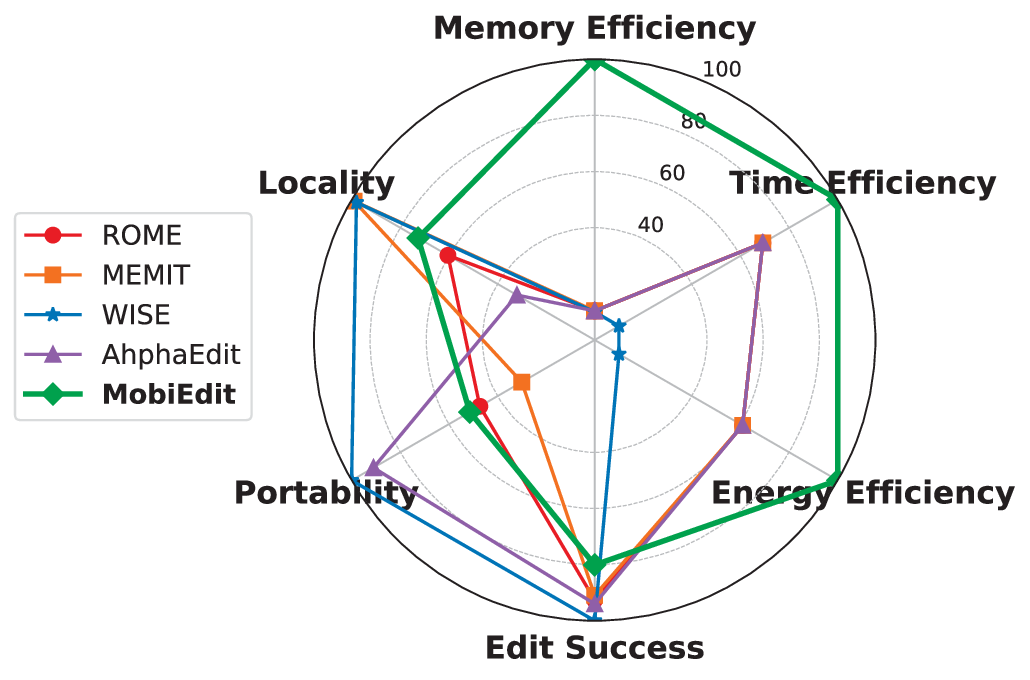}
        \caption*{(a) ZsRE Dataset}
    \end{minipage}
    \hfill
    \begin{minipage}[t]{0.37\linewidth}
        \centering
        \includegraphics[width=\linewidth]{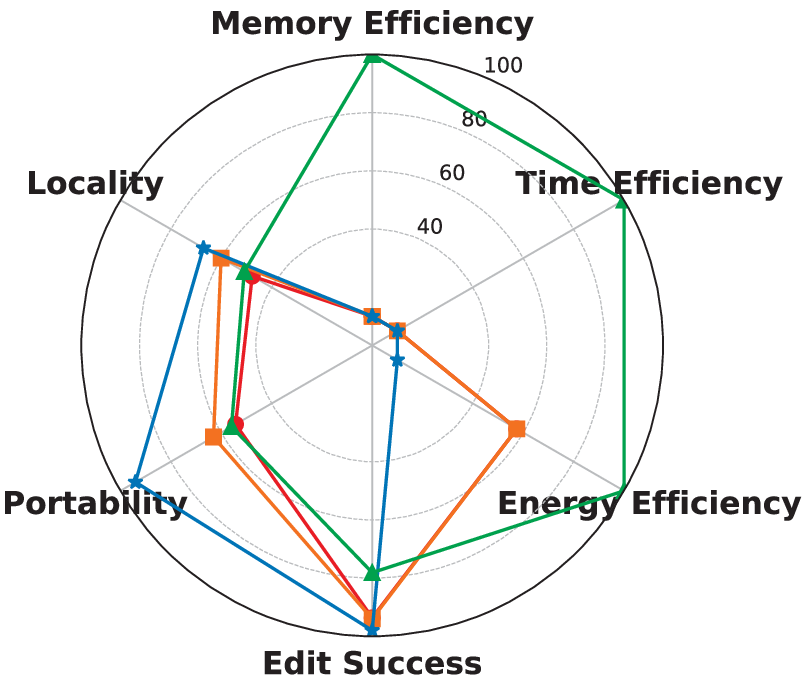}
        \caption*{(b) CounterFact Dataset}
    \end{minipage}
    
    \caption{The comprehensive performance comparison of knowledge editing methods on the ZsRE and CounterFact datasets. 
    System efficieny is obtained as the average of three devices.
    }
    \label{fig:radar_comparison}
\end{figure}

\textbf{Editing performance.}
We compare \sys against all baselines across six dimensions: edit success, locality, portability, time efficiency, memory efficiency, and energy efficiency, as shown in Figure~\ref{fig:radar_comparison}, showing that \textbf{\sys achieves a balance of high accuracy and low resource cost.}

\sys achieves an 80.1 edit success score, 72.6 locality score, and 51.4 portability score while requiring only 25.5 minutes, 0.018 kJ energy, and 6.2 GB memory. 
Although slightly behind in quality (13.9 reduction in edit success loss compared to MEMIT and AlphaEdit), \sys significantly outperforms in efficiency, reducing memory usage by more than 7$\times$ and energy consumption by over 10$\times$. These substantial resource savings make \sys the only realistically viable solution for mobile devices.
The significant performance improvement is attributed to \sys leveraging a bp-free editing method, which reduces the memory and computational overhead caused by backpropagation. Additionally, \sys incorporates mobile hardware-friendly quantization and two optimizations specifically designed for the bp-free training step. These enhancements not only maximize the performance potential of mobile NPUs but also minimize the number of training steps and eliminate redundant computations.
The reduction in accuracy occurs because \sys modifies only the value vector of a single MLP layer, a more resource-efficient and lightweight editing approach compared to the multi-layer editing methods employed by MEMIT and AlphaEdit.

\begin{table}[ht]
\caption{Performance comparison of our method with NPU and other knowledge editing methods with CPU on different devices.}
\centering
\scriptsize
\setlength{\tabcolsep}{3pt}
\renewcommand{\arraystretch}{1.3}
\begin{subtable}[t]{\linewidth}
\caption{ZsRE Dataset}
\begin{tabular}{l c cc cc cc}
\toprule
\multirow{2}{*}{\textbf{Method}} & \multirow{2}{*}{\textbf{Memory (GB)}} 
& \multicolumn{2}{c}{\textbf{K60}} 
& \multicolumn{2}{c}{\textbf{K70}} 
& \multicolumn{2}{c}{\textbf{OnePlus}} \\
& & Time (s) & Energy (J) & Time (s) & Energy (J) & Time (s) & Energy (J) \\
\midrule
ROME & \cellbar{46.14}{0.75} & \cellbar{4543.78}{0.36} & \cellbar{0.25}{0.38} & \cellbar{4276.49}{0.50} & \cellbar{0.24}{0.47} & \cellbar{3252.81}{0.50} & \cellbar{0.18}{0.47} \\
MEMIT & \cellbar{46.14}{0.75} & \cellbar{4543.78}{0.36} & \cellbar{0.25}{0.38} & \cellbar{4276.49}{0.50} & \cellbar{0.24}{0.47} & \cellbar{3252.81}{0.50} & \cellbar{0.18}{0.47} \\
WISE & \cellbar{46.30}{0.77} & \cellbar{11359.44}{0.90} & \cellbar{0.63}{1.00} & \cellbar{8552.99}{1.00} & \cellbar{0.47}{1.00} & \cellbar{6505.63}{1.00} & \cellbar{0.36}{1.00} \\
AhphaEdit & \cellbar{46.14}{0.75} & \cellbar{4543.78}{0.36} & \cellbar{0.25}{0.38} & \cellbar{4276.49}{0.50} & \cellbar{0.24}{0.47} & \cellbar{3252.81}{0.50} & \cellbar{0.18}{0.47} \\
\textbf{MobiEdit} & \cellbargreen{6.20}{0.10} & \cellbargreen{1902.88}{0.18} & \cellbargreen{0.023}{0.05} & \cellbargreen{1477.67}{0.17} & \cellbargreen{0.018}{0.04} & \cellbargreen{1211.83}{0.19} & \cellbargreen{0.014}{0.03} \\
\bottomrule
\end{tabular}
\label{tab:table_zsre}
\end{subtable}
\begin{subtable}[t]{\linewidth}
\caption{CounterFact Dataset}
\begin{tabular}{l c cc cc cc}
\toprule
\multirow{2}{*}{\textbf{Method}} & \multirow{2}{*}{\textbf{Memory (GB)}}
& \multicolumn{2}{c}{\textbf{K60}} 
& \multicolumn{2}{c}{\textbf{K70}} 
& \multicolumn{2}{c}{\textbf{OnePlus}} \\
& & Time (s) & Energy (J) & Time (s) & Energy (J) & Time (s) & Energy (J) \\
\midrule
ROME & \cellbar{46.14}{0.74} & \cellbar{4416.66}{0.23} & \cellbar{0.24}{0.36} & \cellbar{4156.86}{0.25} & \cellbar{0.23}{0.46} & \cellbar{3161.82}{0.26} & \cellbar{0.17}{0.45} \\
MEMIT & \cellbar{46.14}{0.74} & \cellbar{4416.66}{0.23} & \cellbar{0.24}{0.36} & \cellbar{4156.86}{0.25} & \cellbar{0.23}{0.46} & \cellbar{3161.82}{0.26} & \cellbar{0.17}{0.45} \\
WISE & \cellbar{46.30}{0.77} & \cellbar{11041.65}{1.00} & \cellbar{0.61}{1.00} & \cellbar{8313.72}{1.00} & \cellbar{0.46}{1.00} & \cellbar{6323.63}{1.00} & \cellbar{0.35}{1.00} \\
AhphaEdit & \cellbar{46.14}{0.74} & \cellbar{4416.66}{0.23} & \cellbar{0.24}{0.36} & \cellbar{4156.86}{0.25} & \cellbar{0.23}{0.46} & \cellbar{3161.82}{0.26} & \cellbar{0.17}{0.45} \\
\textbf{MobiEdit} & \cellbargreen{6.20}{0.10} & \cellbargreen{1983.17}{0.10} & \cellbargreen{0.024}{0.04} & \cellbargreen{1546.77}{0.08} & \cellbargreen{0.019}{0.02} & \cellbargreen{1271.71}{0.08} & \cellbargreen{0.016}{0.02} \\
\bottomrule
\end{tabular}
\label{tab:table_cf}
\end{subtable}
\label{tab:table_performance}
\end{table}

\textbf{System cost.} 
Table~\ref{tab:table_performance} provides a detailed comparison of memory, latency, and energy usage across three commercial smartphones on ZsRE and CounterFact datasets. All baseline methods, including ROME, MEMIT, AlphaEdit, and WISE, demand excessive resources—over 46GB of memory because the \emph{llm.c} lacks memory optimization on training part of parameters. And their per-edit energy is ranging from 0.18J to 0.63J. For instance, WISE consumes 0.63J and takes 11,359 seconds on K60 for a single edit. Such workloads not only exceed memory budgets but impose intense thermal and scheduling pressure on mobile hardware.

In practical deployment, this level of energy usage often leads to SoC thermal throttling, causing time-dependent slowdowns and inconsistent performance. Moreover, resource contention causes foreground processes (like UI response or system services) to stall during editing tasks lasting over 1.5 to 3 hours, effectively rendering the device unusable during editing.

\sys consumes only 6.2GB memory under 0.03J energy per edit across all devices, completing edits in 1200 to 2000 seconds. This 10$\times$ energy reduction allows \sys to run editing workloads unobtrusively in the background without interrupting the user experience or triggering thermal limits. Such sustainability is critical for real-world mobile applications, where knowledge editing may be triggered interactively under tight system constraints.

\subsection{Ablation Studies}
\begin{wrapfigure}[15]{r}{0.45\textwidth}
  \centering
  \includegraphics[width=0.4\textwidth]{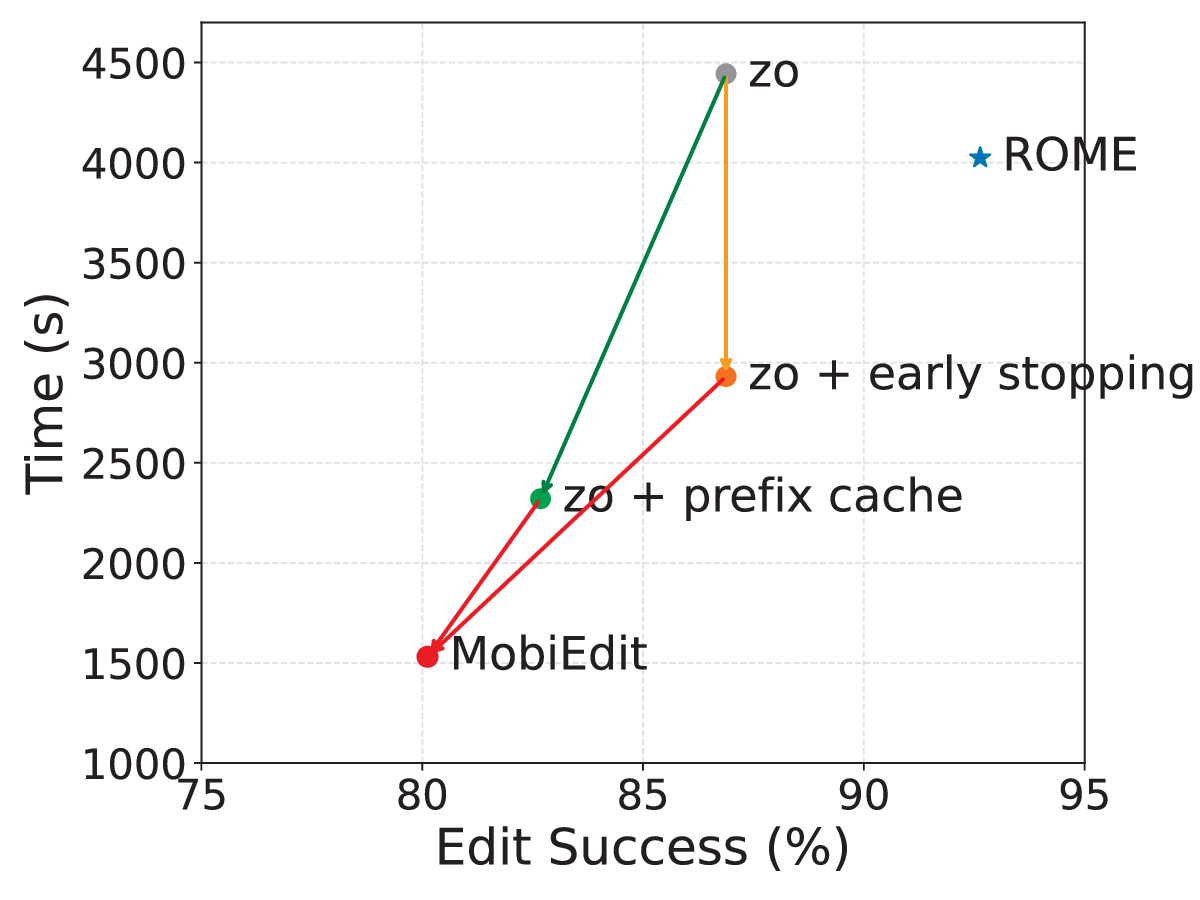}
  \caption{Edit success vs. time on ZsRE. Time is averaged across all devices.}
  \label{fig:ablation}
\end{wrapfigure}

Figure 6 presents an ablation study of the key algorithmic and system-level optimizations in our framework. 
The basic zeroth-order method (zo) achieves moderate edit success but incurs excessive time cost, often over 4000 seconds per edit.
Introducing early stopping (dynamic step controller) alone reduces average editing time by over 40\%, without sacrificing accuracy. The early stopping module effectively eliminates redundant optimization once the target knowledge has been successfully fitted. 
Adding prefix cache further accelerates editing by another 20–30\%, as observed across all devices. For each fact, the prefix cache reduces computation proportionally to the ratio of the prefix length to the total input length.
\sys, which incorporates both optimizations, reduces editing time to nearly one-third of the baseline zo and achieves the best balance of edit success and efficiency.
By leveraging quantization and NPUs, our approach significantly reduces memory usage and editing time by 7$\times$ and 10$\times$, with only a slight reduction in editing success rate. This enables efficient and practical deployment on commercial mobile devices.
This significant performance improvement mainly stems from the usage of floating-point calculations exclusively for a small number (less than 1\%) zhof accuracy-sensitive editing parameters, while the majority of non-trainable parameters are efficiently processed using low-precision integer computations on the mobile NPU.
\sys outperforms the ROME baseline (CPU-based) in terms of speed, underscoring the advantages of quantized, forward-only editing on mobile hardware.
These results highlight the necessity and complementarity of each system-level optimization in achieving practical on-device knowledge editing, delivering substantial improvements in both hardware and algorithmic efficiency.





\section{Related Work}

\subsection{Knowledge Editing Methods}
Recent work on knowledge editing in language models spans several paradigms. Locate-and-editing methods like ROME\cite{meng2022locating} and MEMIT\cite{meng2023mass} directly modify internal weights to update factual knowledge, offering high precision. 
ROME\cite{meng2022locating} operates by identifying a “critical layer” and modifying a specific activation vector, achieving high accuracy in single-fact edits with minimal disruption to the model's behavior.
MEMIT\cite{meng2023mass} extends ROME to support the simultaneous editing of multiple factual associations by modifying activations across several layers.
AlphaEdit\cite{fang2024alphaedit}, also based on ROME and MEMIT, introduces null space projection techniques to preserve the model’s original knowledge during editing.
WISE\cite{wang2024wise} augments the FFN module with a dynamic routing mechanism to accommodate new factual knowledge.In contrast, RAG-based methods like RECIPE\cite{chen-etal-2024-lifelong} avoid modifying model parameters by retrieving external knowledge or prompts, allowing efficient and continual updates. 
MeLLo\cite{zhong_mquake_2023} reveals limitations in generalization under multi-hop reasoning, a challenge for both direct and retrieval-based methods. 
Finally, meta-learning approaches like MEND\cite{mitchell2021fast} learn how to edit from examples, striking a balance between flexibility and generalization but requiring higher setup costs. These approaches trade off between precision, efficiency, and scalability in different ways.

While highly effective in terms of edit success and locality, these approaches depend on multi-step backpropagation, resulting in substantial memory and latency overhead. 
This makes such methods fundamentally incompatible with mobile deployment, where memory is typically limited to 16GB or less, and compute is constrained to forward-only inference on NPUs. In light of these limitations, we focus on memory-efficient, NPU-friendly, and compatible with quantization editing methods that better align with the resource constraints of mobile hardware.

\subsection{BP-Free Methods}

Zero-order optimization (ZO) techniques have gained renewed interest as scalable alternatives to backpropagation, particularly in scenarios where gradient computation is expensive. In the context of model tuning, methods such as FwdLLM~\cite{xu2024fwdllm} and MeZO~\cite{mezo} demonstrate that high-quality adaptation can be achieved using purely forward computations, typically through numerical estimation of gradient signals via perturbation. However, they primarily focus on downstream task adaptation, often requiring thousands of adaptation steps, and have not been applied to factual memory injection. 


To the best of our knowledge, no prior work enables reliable factual editing under both forward-only and quantized constraints. Our method addresses this gap by introducing a backpropagation-free, quantization-aware framework designed specifically for edge deployment.

\section{Conclusion}
We have introduced \sys, the first mobile-compatible framework for efficient knowledge editing in large language models. By replacing backpropagation with a quantized, forward-only gradient estimation technique, \sys aligns with the compute and memory constraints of commercial mobile NPUs. Two further optimizations—prefix cache and early stopping—enhance editing efficiency without sacrificing edit quality. Experimental results demonstrate that \sys enables real-time editing of Qwen2.5-3B entirely on-device, reducing memory usage by 7.6×, energy consumption by 14.7×, and latency by 72\% compared to prior methods. These results highlight \sys as a promising step toward practical, user-driven LLM personalization in mobile and edge scenarios.

\section{Limitations}
\label{sec:limitations}
While \sys achieves efficient on-device knowledge editing, it has two main limitations. First, its edit success is lower than gradient-based methods, especially for more challenging or ambiguous knowledge that lacks clear signal from the target output. Second, \sys currently supports edits in the form of simple subject–object factual pairs. Complex prompts involving multi-hop reasoning, conditional logic, or indirect references are beyond the scope of the current framework.

\begin{ack}
Use unnumbered first level headings for the acknowledgments. All acknowledgments
go at the end of the paper before the list of references. Moreover, you are required to declare
funding (financial activities supporting the submitted work) and competing interests (related financial activities outside the submitted work).
More information about this disclosure can be found at: \url{https://neurips.cc/Conferences/2025/PaperInformation/FundingDisclosure}.

Do {\bf not} include this section in the anonymized submission, only in the final paper. You can use the \texttt{ack} environment provided in the style file to automatically hide this section in the anonymized submission.
\end{ack}

\bibliographystyle{plainnat} 
\bibliography{sample}










\end{document}